\documentclass[runningheads]{llncs}
\usepackage{eccv}

\usepackage{eccvabbrv}

\usepackage{xcolor}
\usepackage{colortbl}
\definecolor{panelgray}{HTML}{EAEAF2}
\usepackage{graphicx}
\usepackage{booktabs}
\usepackage{amsmath,amssymb}
\usepackage{enumitem}
\usepackage{algorithm}
\newcommand{\subfloat}[2][]{\subcaptionbox{#1}{#2}}

\usepackage{caption}

\usepackage[breaklinks,colorlinks,citecolor=eccvblue]{hyperref}

\newcommand{\ensuretext}[1]{#1}
\newcommand{\marker}[2]{\ensuremath{^{\textsc{#1}}_{\textsc{#2}}}}
\newcommand{\arkcomment}[3]{\ensuretext{\textcolor{#3}{[#1 #2]}}}
\newcommand{\wchai}[1]{\arkcomment{\marker{W.H.}{Chai}}{#1}{blue}}

\newif\ifdraft
\draftfalse
\ifdraft
\newcommand{\blc}[1]{{\color{brown}[\textbf{Baiang:} \textit{#1}]}}

\newcommand{\bl}[1]{{\color{purple}#1}}
\else
\newcommand{\blc}[1]{}
\renewcommand{\wchai}[1]{}

\newcommand{\bl}[1]{{\color{black}#1}}
\fi

\title{Weak-to-Strong Knowledge Distillation Accelerates Visual Learning}
\titlerunning{Weak-to-Strong Distillation Accelerates Visual Learning}
\author{
Baiang Li \and
Wenhao Chai \and
Felix Heide
}
\authorrunning{B. Li et al.}
\institute{
Princeton University\\
\email{baiang.li@princeton.edu, wenhao.chai@princeton.edu, fheide@princeton.edu}
}

\begin{document}
\maketitle
\begin{abstract}
    Large-scale visual learning is increasingly limited by training cost.
    Existing knowledge distillation methods transfer from a stronger teacher to a weaker student for compression or final-accuracy improvement.
    We instead investigate distillation to accelerate the training of strong students. We propose a generalizable plug-and-play recipe that freezes a weaker teacher, applies distillation only in early training, and turns it off once the student reaches and surpasses teacher-level performance. For ImageNet and CIFAR classification, this strategy reaches \bl{target thresholds} much earlier, with up to $4.8\times$ speedup measured by epochs.
    We confirm that the method generalizes to other tasks and report $1.7\times$ epoch speedup for object detection on the \bl{COCO dataset}, and $2.5\times$ earlier target-FID crossing for diffusion generation on the \bl{CIFAR-10 dataset}, measured in steps. These findings validate our method as a universal speedup mechanism for visual learning. Project page: \url{https://princeton-computational-imaging.github.io/WeaktoStrong}.
    \end{abstract}

\section{Introduction}
\label{sec:intro}
Modern visual learning increasingly relies on large-scale pretraining to obtain representations that transfer across tasks and domains.
These training recipes typically pair high-capacity models with long schedules and heavy augmentation, making them costly and time-consuming to train.
Moreover, in practice, new model development is often iterative: stronger vision models are frequently built from prior checkpoints, as seen in modern foundation-model lines such as DINOv2~\cite{Oquab2023DINOv2}, EVA-02~\cite{Fang2023EVA02}, SigLIP~\cite{Zhai2023SigLIP}, PaLI-3~\cite{Chen2023PaLI3}, and Florence-2~\cite{Yin2023Florence2}.
We find this creates a concrete opportunity to reuse existing models as training-time guidance and reduce the number of epochs needed to reach \bl{target quality}.

Knowledge Distillation (KD) methods transfer supervision from a teacher to a student and are widely used for model compression and deployment efficiency~\cite{Hinton2015KD,Romero2015FitNets,Park2019RelKD}. Typically, a stronger teacher guides a smaller student throughout training so that the student approaches the teacher’s behavior while retaining much lower inference cost. Beyond compression, KD is also used for regularization and self-distillation~\cite{Furlanello2018BAN,Zhang2019BYOT,Yun2020SelfKD}. However, existing KD approaches are mainly designed for final model quality or inference efficiency, not for minimizing training epochs to a target performance level.

In this work, we investigate weak-to-strong knowledge distillation to accelerate training itself, rather than only for model compression.
Specifically, we measure acceleration by the number of \bl{epochs/steps} needed to reach \bl{target thresholds}.
We find that existing strong-teacher KD often gives limited acceleration, and instead propose a weak-to-strong setup. Our training recipe is intentionally simple: the teacher is weaker than the student, trained once and then frozen; distillation is applied only in the early phase with a short warmup--hold--decay schedule; and distillation is turned off after the student surpasses teacher-level performance. We find that ``suitably weaker'' teachers lie in a moderate gap below the target student (\bl{up to 15\% weaker}), while much weaker or stronger teachers are mismatch settings for acceleration.

We validate the method for classification, image generation, and detection, and report acceleration with the proposed method across all tasks. On ImageNet and CIFAR-10/100 (Table~\ref{tab:cls_results_v2}), our method consistently shifts first@\,$\tau$ \bl{(where $\tau$ denotes the target metric threshold)} milestones earlier for both ResNet and ViT students under SGD, AdamW, and Muon, while keeping final Top-1 close to baseline. Specifically, on ImageNet we observe up to 4.8$\times$ epoch speedup. For object detection on the \bl{COCO dataset} (train5k/val500 split) with RetinaNet-R34$\rightarrow$RetinaNet-R50, we observe about 1.7$\times$ epoch speedup to target AP50; for diffusion generation on the \bl{CIFAR-10 dataset} with nc64-rb2$\rightarrow$nc128-rb3, we observe 2.5$\times$ earlier target-FID crossing measured by training steps. As such, we confirm the method generalizes across visual learning tasks with weak teachers often conveniently available.

These observations motivate the following contributions.
\begin{itemize}
    \item \textit{Plug-and-Play Training Acceleration.} We propose a simple weak-to-strong distillation recipe that adds early-phase distillation with warmup--hold--decay and stops distillation after teacher-level surpass, while keeping the baseline student training pipeline unchanged.
    \item \textit{Analysis of Acceleration Regime}. We characterize when acceleration is reliable, showing that suitably weaker teachers in a moderate gap below the student provide the strongest gains, while too-weak or too-strong teachers reduce gains; this pattern is consistent across SGD, AdamW, and Muon.
    \item \textit{Validation Across Diverse Visual Learning Tasks.} We validate transfer across image classification, object detection, and diffusion generation, with up to 4.8$\times$ speedup on ImageNet classification, 1.7$\times$ for detection on the \bl{COCO dataset}, and more than 2.5$\times$ for diffusion generation on the \bl{CIFAR-10 dataset}.

\end{itemize}

\begin{figure*}[t]
  \centering
  \includegraphics[width=\linewidth]{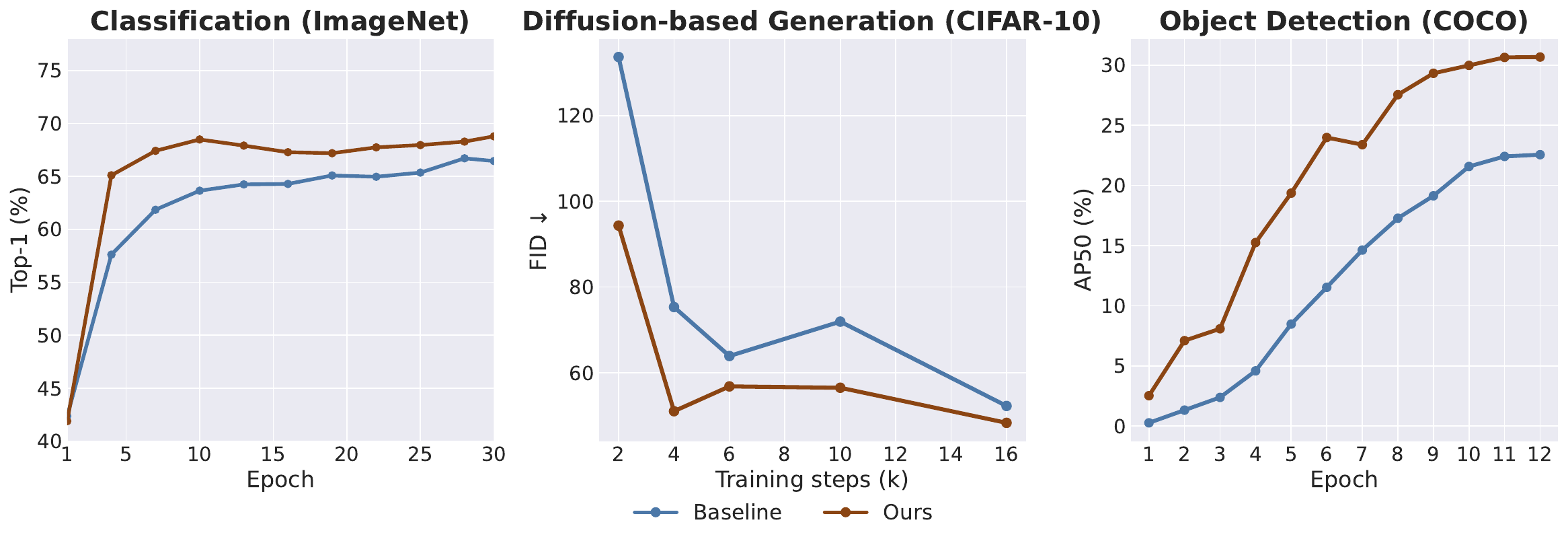}
  \caption{\textbf{Weak-to-\bl{Strong} Distillation Accelerates Visual Learning.} Blue curves are baseline training and red curves are with our method. From left to right: classification on ImageNet, diffusion-\bl{b}ased generation on the \bl{CIFAR-10 dataset}, and object detection on the \bl{COCO dataset}. Our method reaches target quality earlier in all three tasks: 65\% Top-1 on ImageNet (4.8$\times$ fewer epochs), target FID on diffusion (more than 2.5$\times$ fewer training steps), and target AP50 for object detection on the \bl{COCO dataset} (1.7$\times$ fewer epochs).
}
  \label{fig:teacher_range}
\end{figure*}

\section{Related Work}
\label{sec:related}
\paragraph{Training Acceleration.}
Training acceleration in visual learning is commonly pursued through optimizer and recipe design, including SGD with momentum~\cite{RobbinsMonro1951SGD,Polyak1964Momentum}, AdamW~\cite{Loshchilov2019AdamW}, and Muon~\cite{Jordan2024Muon}, together with modern training pipelines~\cite{Wightman2021ResNetStrikesBack}.
Augmentation and regularization policies such as Mixup~\cite{Zhang2018Mixup}, CutMix~\cite{Yun2019CutMix}, RandAugment~\cite{Cubuk2020RandAugment}, and Random Erasing~\cite{Zhong2020RandomErasing} also have been proposed to improve convergence quality and stability.
These strategies are highly effective, but have not investigated teacher-student transfer to directly shorten target-reaching epochs. Distillation-oriented acceleration lines, such as KDEP~\cite{He2022KDEP}, FKD~\cite{Shen2022FKD}, DearKD~\cite{Chen2022DearKD}, and CSKD~\cite{Zhao2023CSKD}, further highlight the importance of scheduling and data efficiency for faster convergence. Our work is complementary: we keep the baseline optimizer and data recipe fixed, and propose our weak-to-strong method as a plug-in acceleration mechanism.

\paragraph{Knowledge Distillation.}
Traditional KD transfers a trained teacher's knowledge to a student by matching softened logits or intermediate representations.
Early formulations include temperature-scaled logit KD~\cite{Hinton2015KD}, intermediate hint supervision in FitNets~\cite{Romero2015FitNets}, and relation- or contrastive-based objectives that preserve inter-sample structure~\cite{Park2019RelKD,Tian2020CRD}.
Recent work refines the loss (e.g., DKD~\cite{Zhao2022DKD}) or adds task-aware signals such as multi-level logits~\cite{Song2023MultiModeKD}, class attention~\cite{Guo2023ClassAttentionKD}, and pixel-level contrast~\cite{Huang2023PixelContrast}; specialized variants target long-tailed ViTs~\cite{Rangwani2024DeiTLT}, self-supervised learning~\cite{Song2023MultiModeKD,Wang2023MaskedVideoKD}, and diffusion models~\cite{Meng2023GuidedDiffusionKD,Hsiao2024PlugPlayKD,Koo2024PosteriorDistillation}.
Most of these methods assume a stronger teacher and optimize final accuracy or compression; far fewer works study how KD affects time-to-target \bl{metrics} under matched compute and training recipes.
A line of data-efficient and early-phase KD studies (e.g., DeiT~\cite{Touvron2021DeiT}, DearKD~\cite{Chen2022DearKD}, CSKD~\cite{Zhao2023CSKD}, KDEP~\cite{He2022KDEP}, FKD~\cite{Shen2022FKD}, patient/consistent teachers~\cite{Beyer2022PatientKD}, and theory on variance reduction~\cite{Safaryan2023VarianceKD}) highlights the role of scheduling.
We focus explicitly on training speed: we use suitably weak teachers and stop distillation once the student surpasses the teacher to accelerate early milestones while preserving final accuracy under matched training recipes.

\paragraph{Early Distillation.}
A recurring observation in KD methods is that teacher guidance is most beneficial in early optimization and should be weakened or stopped later to avoid redundant constraints near convergence. DeiT explores data-efficient distillation in practical vision training~\cite{Touvron2021DeiT}, while DearKD and CSKD emphasize schedule design for better transfer dynamics~\cite{Chen2022DearKD,Zhao2023CSKD}. KDEP and FKD further improve efficiency by reducing teacher-forward overhead through early-phase or cached distillation pipelines~\cite{He2022KDEP,Shen2022FKD}, and patient-teacher analyses and theory connect these effects to optimization stability and variance reduction~\cite{Beyer2022PatientKD,Safaryan2023VarianceKD}. The proposed method also finds early transfer most intuitive but differs in objective and protocol: instead of seeking better final accuracy with a stronger teacher, we use a deliberately weaker frozen teacher and a surpass-based stopping rule to improve \bl{first@\,$\tau$} speedup at \bl{target thresholds}.
\vspace*{-8pt}
\paragraph{Vision Pretraining.}
Foundation-scale vision models increasingly rely on long-horizon pretraining that blends self-supervision, language-image alignment, and promptable segmentation. Recent work such as DINOv2~\cite{Oquab2023DINOv2}, EVA-02~\cite{Fang2023EVA02}, I-JEPA~\cite{Assran2023IJEPA}, and InternImage~\cite{Wang2023InternImage} scales masked or predictive objectives to large corpora and yields highly transferable features. Vision-language efforts (SigLIP~\cite{Zhai2023SigLIP}, PaLI-3~\cite{Chen2023PaLI3}) and unified decoders (Florence-2~\cite{Yin2023Florence2}) broaden this to multi-task interfaces, while Segment Anything~\cite{Kirillov2023SAM} shows promptable segmentation at scale. Within this landscape, we position our method as a plug-and-play training recipe tool to shorten iterative training runs: pair the student with a suitably weaker pretrained checkpoint, use a short early distillation phase, and stop once the student surpasses the teacher to preserve final accuracy.

\vspace*{-8pt}
\paragraph{Weak-to-\bl{S}trong (\bl{I}nverse) \bl{D}istillation.}
Several works have aimed at inverting the traditional KD hierarchy.
Born-Again Networks grow students iteratively from their predecessors~\cite{Furlanello2018BAN}, while self-distillation approaches such as BYOT and SKD regularize a network using its own softened outputs or auxiliary heads~\cite{Zhang2019BYOT,Yun2020SelfKD}.
Peer-based variants (Deep Mutual Learning~\cite{Zhang2018DML}) and Teacher-Assistant KD~\cite{Mirzadeh2020TAKD} co-train or bridge large capacity gaps.
These methods improve generalization but retain strong (or equal) teachers and usually prioritize final metrics.
Our work instead focuses on a compact, frozen teacher that is deliberately slightly weaker than the target model and is used purely to accelerate early optimization.
This ``weak-to-strong'' perspective complements prior inverse KD work by explicitly targeting training speed.

Weak-to-strong alignment has recently become prominent in large-model and vision literature. The original weak-to-strong study by Burns et al.~\cite{Burns2023WeakStrong} demonstrates that weak supervision can elicit strong capabilities in language models, while Vision Superalignment generalizes the idea to vision foundation models with robustness and transfer objectives~\cite{Guo2024VisionSuperalignment}. Gambashidze et al.\ extend the paradigm to LiDAR 3D detection~\cite{Gambashidze2024W2S3D}. Earlier semi-supervised methods such as Noisy Student~\cite{Xie2020NoisyStudent} and Meta Pseudo Labels~\cite{Pham2021MetaPseudo} also show that weaker teachers can bootstrap stronger students, though their primary goal is final accuracy.
\bl{In contrast, our focus is training acceleration: we evaluate weak-to-strong KD with matched recipes and report first@\,$\tau$ speedups as the primary metric, rather than only final-task performance.}

\section{Weak-to-Strong Distillation}
\label{sec:method}
We list the proposed weak-to-strong training acceleration method in Algorithm~\ref{alg:w2s}. Specifically, our method is a plug-and-play early-training add-on: keep a weaker teacher frozen, add a distillation loss in early training, and turn distillation off after the student reaches teacher-level performance. Algorithm~\ref{alg:w2s} gives a \bl{task-agnostic} loop; only $\mathcal{L}_{\text{base}}$ and $\mathcal{L}_{\text{distill}}$ are \bl{instantiated per task} below (classification, detection, generation).

\paragraph{Training Setup.}
Let $f_\theta$ be a trainable strong student and $g_\phi$ be a frozen weak teacher, where $\theta$ and $\phi$ are student and teacher parameters. We augment the original task loss $\mathcal{L}_{\text{base}}$ with a distillation term $\mathcal{L}_{\text{distill}}$ as
\begin{equation}
\label{eq:loss}
\mathcal{L}(u) = \mathcal{L}_{\text{base}} + \bl{\gamma}\,\lambda(u)\,\mathcal{L}_{\text{distill}}.
\end{equation}
Here, the training index $u$ parametrizes the training stage (epoch/iteration), and  $\lambda(u)\in[0,\lambda_{\max}]$ is the stage-dependent distillation weight. When $\lambda(u)=0$, training is exactly the baseline recipe.
\begin{algorithm}[t]
\footnotesize
\caption{\textbf{Proposed Weak-to-Strong Training.}}
\label{alg:w2s}
\textbf{Input:} frozen teacher $g_\phi$, student $f_\theta$, train data $\mathcal D$, val data $\mathcal D_{\mathrm{val}}$, schedule $\Lambda(t)$, scale $\gamma$, stop length $k$.\\
\textbf{Output:} $\theta^\star$.
\[
s(m,m_{\mathrm{ref}})=
\begin{cases}
\mathbb{1}[m\ge m_{\mathrm{ref}}], & \text{Top-1/AP50},\\
\mathbb{1}[m\le m_{\mathrm{ref}}], & \text{FID}.
\end{cases}
\]
\begin{enumerate}[leftmargin=*,itemsep=0.8pt,topsep=2pt]
  \item $m_{\mathrm{ref}}\!\leftarrow\!\mathrm{Eval}(g_\phi,\mathcal D_{\mathrm{val}}),\; c\!\leftarrow\!0,\; a\!\leftarrow\!1$.
  \item For $t=1,\ldots,T$: $\lambda_t\!\leftarrow\!a\,\Lambda(t)$.
  \item For $(x,y)\in\mathcal D$:
  \[
  L_{\mathrm{base}}=\mathcal L_{\mathrm{base}}^{\text{task}}(x,y),\quad
  L_{\mathrm{distill}}=\mathcal L_{\mathrm{distill}}^{\text{task}}(x,y;g_\phi,f_\theta),
  \]
  \[
  \theta\leftarrow \mathrm{OptStep}\!\left(\theta,\nabla_\theta\!\big[L_{\mathrm{base}}+\gamma\lambda_t L_{\mathrm{distill}}\big]\right).
  \]
  \item $m_t\leftarrow\mathrm{Eval}(f_\theta,\mathcal D_{\mathrm{val}}),\;\; c\leftarrow s(m_t,m_{\mathrm{ref}})\,(c+1)$.
  \item If $c\ge k$, set $a\leftarrow0$ (thus $\lambda_{t'>t}=0$).
  \item Return $\theta^\star\leftarrow\theta$.
\end{enumerate}
\end{algorithm}

\paragraph{Early Training Schedule and Stop Rule.} We employ a short warmup-hold-decay schedule for $\lambda(u)$. Warmup prevents unstable gradients at the beginning, hold provides consistent transfer in early training, and decay prevents late-stage over-regularization. We use \bl{an} adaptive stop rule: once the student surpasses the frozen teacher for $k$ consecutive validations, we set $\lambda(u)=0$ permanently (we use $k=2$). Let $m$ denote the validation metric used for stopping. For `higher-is-better metrics' (Top-1, AP50), surpassing means the student validation metric is at least the teacher metric. For `lower-is-better' metrics (FID), surpass means the student validation metric is at most the teacher metric.
Schedule and post-surpass diagnostic plots are provided in the Supplementary Material. 

\paragraph{Weak Teacher Selection.}
We define a weak teacher as a model whose standalone performance is \bl{up to 15\% weaker than the baseline student}. The corresponding operational band is shown in Figure~\ref{fig:teacher_range_v2}.

\vspace{\baselineskip}
In the following, we describe the concrete instantiations for classification, object detection, and image generation.
\paragraph{Classification.}
For image classification, $\mathcal{L}_{\text{base}}$ is standard cross-entropy with one-hot or label-smoothed targets, that is
\begin{equation}
\label{eq:loss_ce}
\mathcal{L}_{\text{base}}(x,y)= -\sum_{c=1}^{K} q_c(y)\log p_{s,c}(x),\quad p_s(x)=\mathrm{softmax}(f_\theta(x)).
\end{equation}
Here $x$ is an input image, $y$ is its label, $K$ is the number of classes, $q_c(y)$ is the target distribution (one-hot or label-smoothed), and $p_{s,c}(x)$ is the student probability of class $c$. Distillation uses forward KL between teacher and student softened posteriors
\begin{equation}
\label{eq:loss_kd}
\mathcal{L}_{\text{distill}}(x,u)=\mathcal{T}(u)^2\,\mathrm{KL}\!\left(p_t^{\mathcal{T}(u)}(x)\,\|\,p_s^{\mathcal{T}(u)}(x)\right),
\end{equation}
where $\mathcal{T}(u)$ is the temperature schedule, and $p_t^{\mathcal{T}}(x)=\mathrm{softmax}(g_\phi(x)/\mathcal{T})$ and $p_s^{\mathcal{T}}(x)=\mathrm{softmax}(f_\theta(x)/\mathcal{T})$. We use forward KL following~\cite{Hinton2015KD}, with temperature decayed from 6 to 1.

\paragraph{Object Detection.}
For object detection, $\mathcal{L}_{\text{base}}$ is the \bl{original detector loss (classification + box regression)}~\cite{Lin2017FocalLoss,Ren2015FasterRCNN}. Our distillation term aligns teacher and student prediction heads on the same images. We distill classification logits with temperature scaling and confidence masking (teacher-score threshold), and optionally add a box-regression alignment term, that is
\[
\bl{\mathcal{L}_{\text{distill}}^{\text{det}} = \mathcal{L}_{\text{cls-distill}} + \beta\,\mathcal{L}_{\text{box-distill}},}
\]
where \bl{$\mathcal{L}_{\text{cls-distill}}$} distills classification logits, \bl{$\mathcal{L}_{\text{box-distill}}$} distills box regression, and $\beta\ge 0$ controls the box term (set to 0 when box distillation is disabled). In all experiments, we keep the \bl{same detector optimizer and data pipeline as baseline} and only add this early-stage distillation schedule.

\paragraph{Generation.}
For diffusion generation, $\mathcal{L}_{\text{base}}$ is the standard denoising objective (MSE plus variational-bound term, VB). The distillation signal aligns teacher and student noise predictions on the same noised sample $(x_t,t,\epsilon)$, where $t$ denotes diffusion timestep (different from the outer training index $u$ in Eq.~\eqref{eq:loss}) as
\[
\mathcal{L}_{\text{distill}}^{\text{gen}}=\left\|\epsilon_\theta(x_t,t)-\epsilon_\phi(x_t,t)\right\|_2^2.
\]
Here $x_t$ is the noised sample at diffusion timestep $t$, $\epsilon\sim\mathcal{N}(0,I)$ is sampled Gaussian noise, and $\epsilon_\theta,\epsilon_\phi$ are student/teacher noise predictions. In our experiments, acceleration is measured by the first training step reaching the teacher-level FID target and reported as baseline/ours step speedup.

\paragraph{What Is Shared and What Is Task-Specific.}
In addition to Algorithm~\ref{alg:w2s}, the method shares across all tasks: frozen weak teacher, early-stage schedule, and stop-after-surpass principle. Task-specific are: the base objective, the distillation signal, and the validation metric used for surpass. We find that this recipe enables consistent acceleration across all tasks assessed in the following.

\section{Experiments}
\label{sec:experiments_v2}
Before reporting our evaluation results, we describe the experimental setting and implementation details. 
\subsection{Experimental Setting}
\label{sec:exp_v2_setup}
For each experiment, unless explicitly noted otherwise, we keep the same student architecture, optimizer, data pipeline, and training budget, and only add our early-stage distillation module. For each evaluation, we record \bl{first@\,$\tau$} for baseline and ours, \bl{where $\tau$ denotes the target metric threshold}, then report speedup ratio as baseline \bl{first@\,$\tau$} divided by ours \bl{first@\,$\tau$}. A speedup ratio greater than $1.0\times$ means ours reaches the same target earlier with fewer epochs or steps. For higher-is-better metrics (Top-1, AP50), \bl{first@\,$\tau$} is the first epoch with metric at or above $\tau$. For lower-is-better metrics (FID), \bl{first@\,$\tau$} is the first step at or below $\tau$.

\paragraph{Gate and Hyperparameter Selection.} For ImageNet classification~\cite{Deng2009ImageNet}, we use $\tau=65$ for ResNet-50~\cite{He2016ResNet} and $\tau=50$ for ViT-S/16~\cite{Dosovitskiy2021ViT}. For CIFAR early-stage classification~\cite{Krizhevsky2009CIFAR}, we use fixed dataset-level gates: $\tau=75$ for CIFAR-10 and $\tau=60$ for CIFAR-100. For object detection on the \bl{COCO dataset}~\cite{Lin2014COCO}, we use a \bl{fixed task-level} AP50 target ($\tau=20\%$). For diffusion generation on the \bl{CIFAR-10 dataset}~\cite{Krizhevsky2009CIFAR,Ho2020DDPM,Nichol2021ImprovedDDPM}, we use a \bl{fixed task-level FID target} ($\tau=60$), selected from the teacher reference around 30k training steps, with a conservative consecutive-hit rule $k=2$. For classification, we use temperature decay 6$\rightarrow$1 with a warmup-hold-decay weight schedule. In detection, the reported run uses \bl{focal-logit distillation}~\cite{Lin2017FocalLoss,Hinton2015KD}, with $T=2$, score threshold 0.2, bbox distillation weight 1.0, and distillation scale $\lambda=2.0$. In diffusion generation, the reported run uses distillation weight 0.2 with early-step timestep masking ratio 0.5.

\subsubsection{Image Classification Task}
\label{sec:exp_v2_setup_cls}
For image classification, we use ImageNet-1K~\cite{Deng2009ImageNet} and CIFAR-10/100~\cite{Krizhevsky2009CIFAR}, choose teacher-student pairs including ResNet-18$\rightarrow$ResNet-50~\cite{He2016ResNet}, MobileNetV2$\rightarrow$ResNet-50 and MobileNetV2$\rightarrow$ViT-S/16~\cite{Sandler2018MobileNetV2,Dosovitskiy2021ViT}, and DenseNet-40$\rightarrow$ResNet-18/DenseNet-100~\cite{Huang2017DenseNet,He2016ResNet}, and evaluate with Top-1, \bl{first@\,$\tau$} epoch, and speedup ratio (Baseline/Ours).

\subsubsection{Object Detection Task}
\label{sec:exp_v2_setup_det}
For \bl{object} detection on the \bl{COCO dataset}, we use the train5k/val500 split~\cite{Lin2014COCO}, evaluate RetinaNet-R34$\rightarrow$RetinaNet-R50~\cite{Lin2017FocalLoss} and Faster R-CNN-R18$\rightarrow$Faster R-CNN-R50~\cite{Ren2015FasterRCNN}, and report AP50, \bl{first@\,$\tau$} epoch, and speedup ratio (Baseline/Ours).

\subsubsection{Image Generation Task}
\label{sec:exp_v2_setup_gen}
For image generation \bl{on the CIFAR-10 dataset}, we evaluate diffusion~\cite{Krizhevsky2009CIFAR,Ho2020DDPM,Nichol2021ImprovedDDPM}, choose a smaller DDPM-style U-Net teacher (nc64-rb2) and a larger DDPM-style U-Net student (nc128-rb3)~\cite{Ho2020DDPM,Nichol2021ImprovedDDPM}, and report FID~\cite{Heusel2017FID}, \bl{first@\,$\tau$} step, and speedup ratio (Baseline/Ours).

\paragraph{Teacher Availability Assumption.}
We assume the teacher is available, either as a public checkpoint or from a prior training run. 
During the distillation-active phase, we run a single forward pass of the frozen teacher per mini-batch and optionally \bl{precompute/cache} teacher outputs when the data pipeline is reproducible.
Therefore, main-text tables focus on epoch/step speedup ratios. We additionally report wall-clock accounting when teacher re-training is included in the Supplementary Material.

\subsection{Evaluation}
\label{sec:exp_v2_results}

\subsubsection{Classification Results}
\label{sec:exp_v2_results_cls}

Table~\ref{tab:cls_results_v2} reports time-to-target results on ImageNet~\cite{Deng2009ImageNet} and CIFAR~\cite{Krizhevsky2009CIFAR}. Across the listed settings, ours consistently reaches the same targets earlier or at parity, with the strongest gain on ImageNet Muon (4.75$\times$) and clear gains on CIFAR (up to 1.83$\times$). The classification curves show a stable left-shift: targets are reached earlier while best Top-1 remains close to baseline. This trend is confirmed across multiple teacher--student pairs and optimizers, supporting that our method improves optimization speed rather than only final accuracy. Total training time reduction experiments are reported in the Supplementary Material.

\begin{table*}[t]
\centering
\small
\caption{\textbf{Image Classification Training Acceleration.}
ImageNet and CIFAR-10/100 results across representative teacher--student pairs and optimizers.}
\label{tab:cls_results_v2}
\setlength{\tabcolsep}{4.0pt}
\begingroup
\renewcommand{\arraystretch}{0.96}
\resizebox{\textwidth}{!}{%
\begin{tabular}{>{\columncolor{panelgray}}l|>{\columncolor{panelgray}}l|>{\columncolor{panelgray}}l|>{\columncolor{panelgray}}c|>{\columncolor{panelgray}}c|>{\columncolor{panelgray}}c|>{\columncolor{panelgray}}c|>{\columncolor{panelgray}}c}
\toprule
Dataset & Student & Teacher & Optimizer & Teacher Top-1 (\%) & Target $\tau$ & Speedup & \bl{Best} Top-1 (Base / Ours) \\
\midrule
ImageNet & ResNet-50 & ResNet-18   & SGD   & 70.73 & 65 & 1.62$\times$ & 76.52 / \textbf{76.81} \\
ImageNet & ResNet-50 & ResNet-18   & AdamW & 70.73 & 65 & \textbf{2.86$\times$} & 77.19 / \textbf{77.72} \\
ImageNet & ResNet-50 & ResNet-18   & Muon  & 70.73 & 65 & \textbf{4.75$\times$} & \textbf{77.11} / \textbf{77.11} \\
ImageNet & ResNet-50 & MobileNetV2 & SGD   & 69.62 & 65 & 1.48$\times$ & 76.52 / \textbf{76.77} \\
ImageNet & ResNet-50 & MobileNetV2 & AdamW & 69.62 & 65 & 2.00$\times$ & 77.19 / \textbf{77.56} \\
ImageNet & ResNet-50 & MobileNetV2 & Muon  & 69.62 & 65 & \textbf{3.17$\times$} & \textbf{77.11} / 77.09 \\
ImageNet & ViT-S/16  & MobileNetV2 & AdamW & 69.62 & 50 & 1.45$\times$ & \textbf{75.87} / 75.36 \\
CIFAR-100 & ResNet-18    & DenseNet-40 & SGD   & 69.70 & 60 & 1.60$\times$ & 78.75 / \textbf{79.36} \\
CIFAR-100 & DenseNet-100 & DenseNet-40 & SGD   & 69.70 & 60 & 1.83$\times$ & 82.08 / \textbf{82.51} \\
CIFAR-10  & ResNet-50    & MobileNetV2 & AdamW & 85.55 & \bl{75} & \bl{1.60$\times$} & \bl{93.95 / \textbf{94.00}} \\
CIFAR-10  & ResNet-50    & MobileNetV2 & SGD   & 85.55 & 75 & 1.36$\times$ & \textbf{95.65} / 95.08 \\
\bottomrule
\end{tabular}}
\endgroup
\end{table*}
\vspace{-1.5mm}

\subsubsection{Object Detection Results}
\label{sec:exp_v2_results_det}

Table~\ref{tab:cross_task_results_v2} (upper half) reports object detection on the COCO dataset~\cite{Lin2014COCO} under the same target protocol ($\tau=20\%$ AP50).
Without changing the baseline detector training recipe, the same early weak-to-strong distillation template transfers across detector families: RetinaNet-R34$\rightarrow$RetinaNet-R50~\cite{Lin2017FocalLoss} reaches the target at 10$\rightarrow$6 epochs (1.67$\times$), and Faster R-CNN-R18$\rightarrow$Faster R-CNN-R50~\cite{Ren2015FasterRCNN} reaches the target at 4$\rightarrow$3 epochs (1.33$\times$).
This indicates the acceleration effect is not restricted to image classification losses.

\subsubsection{Generation Results}
\label{sec:exp_v2_results_gen}

Table~\ref{tab:cross_task_results_v2} (lower half) reports diffusion generation on the CIFAR-10 dataset~\cite{Krizhevsky2009CIFAR,Ho2020DDPM,Nichol2021ImprovedDDPM} with a fixed target protocol ($\tau=60$). Across student architecture variants, ours reaches \bl{first@\,$\tau$} earlier in the selected positive cases, with speedups from 1.50$\times$ to 2.67$\times$. For generation, we count target crossing only when the metric stays beyond the threshold for two consecutive evaluations, rather than at the first crossing, which is more reliable for noisy curves. These results confirm that early teacher guidance can reduce optimization steps to enter the target quality region.

\begin{table*}[t]
\centering
\small
\caption{\textbf{Detection and Generation Training Acceleration.}
Results for object detection on the COCO dataset and diffusion generation on the CIFAR-10 dataset across representative teacher--student pairs.}
\label{tab:cross_task_results_v2}
\setlength{\tabcolsep}{4.0pt}
\begingroup
\renewcommand{\arraystretch}{0.96}
\resizebox{\textwidth}{!}{%
\begin{tabular}{>{\columncolor{panelgray}}l|>{\columncolor{panelgray}}l|>{\columncolor{panelgray}}l|>{\columncolor{panelgray}}l|>{\columncolor{panelgray}}c|>{\columncolor{panelgray}}c|>{\columncolor{panelgray}}c|>{\columncolor{panelgray}}c}
\toprule
Task & Dataset & Student & Teacher & Target $\tau$ & \bl{first@\,$\tau$} (Base $\rightarrow$ Ours) & Speedup & Best Metric (Base / Ours) \\
\midrule
Detection & COCO & RetinaNet-R50 & RetinaNet-R34 & AP50 20.0\% & 10 ep $\rightarrow$ \textbf{6 ep} & \textbf{1.67$\times$} & AP50: 22.55 / \textbf{30.67} \\
Detection & COCO & Faster R-CNN-R50 & Faster R-CNN-R18 & AP50 20.0\% & 4 ep $\rightarrow$ \textbf{3 ep} & 1.33$\times$ & AP50: 35.97 / \textbf{36.72} \\
\midrule
Generation & CIFAR-10 & nc128-rb3 & nc64-rb2 & FID 60 & 16k $\rightarrow$ \textbf{6k} & \textbf{2.67$\times$} & FID$\downarrow$: 52.27 / \textbf{47.22} \\
Generation & CIFAR-10 & nc160-rb3 & nc64-rb2 & FID 60 & 18k $\rightarrow$ \textbf{12k} & \textbf{1.50$\times$} & FID$\downarrow$: 53.49 / \textbf{47.67} \\
\bottomrule
\end{tabular}}
\endgroup
\end{table*}
\vspace{-1.5mm}

\subsection{Ablation Experiments}
\label{sec:exp_v2_ablation}

\subsubsection{Switching Teachers and Students.}
\label{sec:exp_v2_ablation_teacher_student}

\begin{figure*}[t]
  \centering
  \subfloat[Keep student fixed, switch teachers.]{\includegraphics[width=0.495\linewidth]{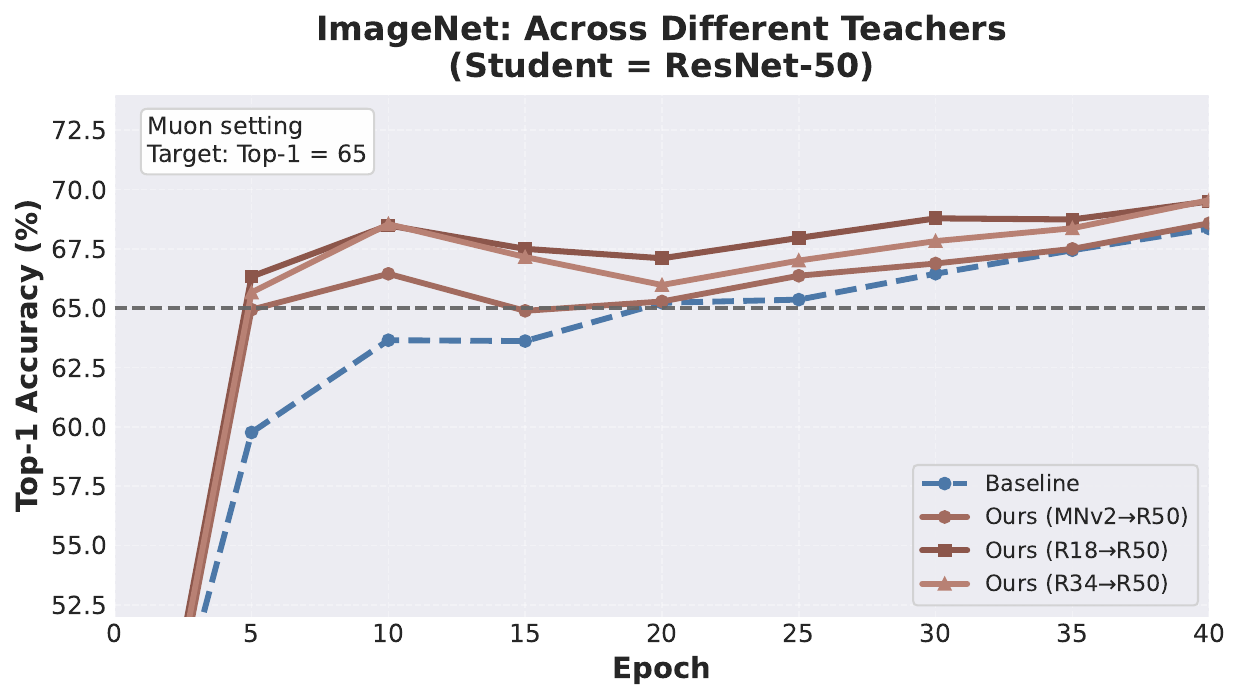}\label{fig:ablation_switch_teacher_v2}}\hfill
  \subfloat[Keep teacher \bl{fixed} (DenseNet-40), and \bl{vary} CIFAR-100 students.]{\includegraphics[width=0.495\linewidth]{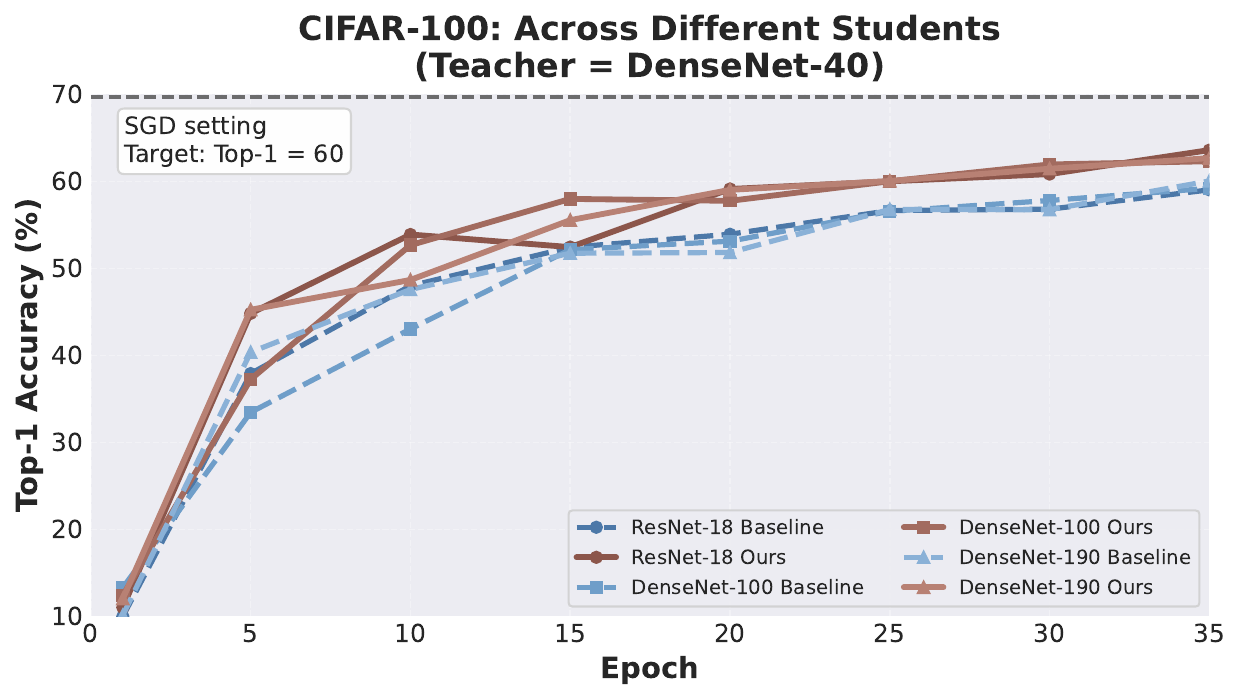}\label{fig:ablation_switch_student_v2}}
  \caption{\textbf{Teacher/Student Swap.}
  (a) ImageNet teacher swap with fixed student; (b) CIFAR-100 student swap with fixed DenseNet-40 teacher. Both panels preserve earlier target-reaching under matched training recipes.}
  \label{fig:ablation_switch_pairs_v2}
\end{figure*}

Figure~\ref{fig:ablation_switch_pairs_v2} reports two complementary experiments on teacher-side and student-side transfer.
Keeping the student fixed and switching teachers on ImageNet shows that suitably weaker teachers keep strong acceleration.
Keeping the teacher fixed and switching CIFAR-100 students with a DenseNet-40 teacher~\cite{Huang2017DenseNet} shows the same early left-shift across student architectures.

\begin{figure*}[t]
  \centering
  \includegraphics[width=0.985\linewidth]{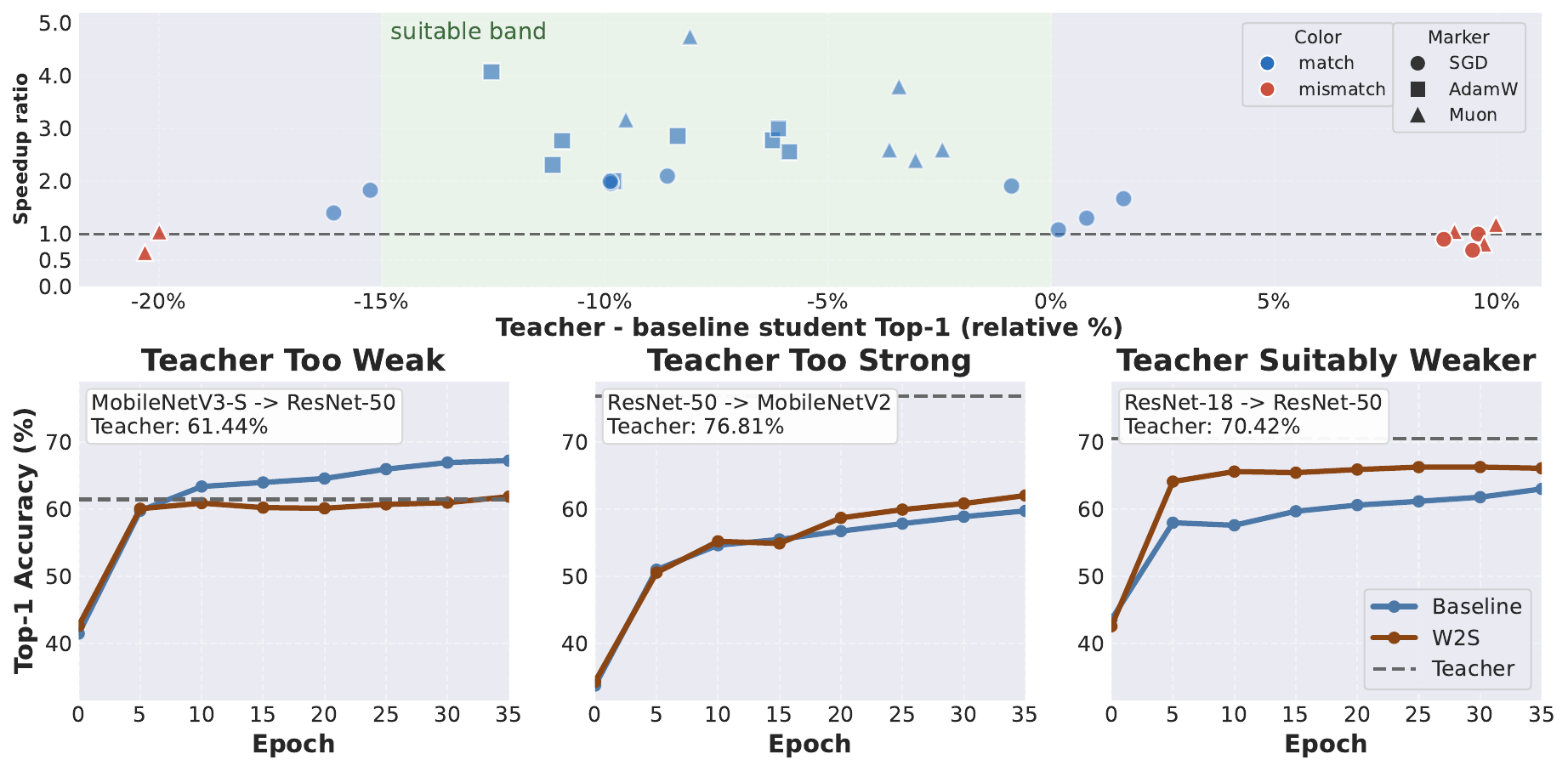}
  \caption{\textbf{Teacher Operational Band.}
  Top: speedup ratio versus relative teacher--student Top-1 gap (teacher minus baseline student, relative \%). Bottom: representative too-weak, too-strong, and suitably-weaker trajectories under matched recipes.}
  \label{fig:teacher_range_v2}
\end{figure*}

Figure~\ref{fig:teacher_range_v2} summarizes teacher suitability.
The top panel maps speedup ratio versus relative teacher--student gap, and the bottom panels already show the three mismatch regimes (too weak, too strong, suitably weaker). The practical operating band is where teacher accuracy is moderately below baseline student accuracy; this is where acceleration is most reliable.

\vspace{-1.5mm}
\begin{table*}[t]
\centering
\small
\caption{\textbf{Teacher Mismatch Across Tasks.}
Too-weak, too-strong, and suitably-weaker teacher settings under matched training recipes; for COCO we keep the same backbone pair (R34$\rightarrow$R50) and only change teacher \bl{checkpoint stage}: \bl{earlier} checkpoint (too weak), \bl{later} checkpoint (too strong), and \bl{mid} checkpoint (suitably weaker).}
\label{tab:mismatch_behavior_v2}
\setlength{\tabcolsep}{4.2pt}
\begingroup
\renewcommand{\arraystretch}{0.96}
\resizebox{\textwidth}{!}{%
\begin{tabular}{>{\columncolor{panelgray}}l|>{\columncolor{panelgray}}l|>{\columncolor{panelgray}}l|>{\columncolor{panelgray}}l|>{\columncolor{panelgray}}c|>{\columncolor{panelgray}}c|>{\columncolor{panelgray}}c}
\toprule
Task & Regime & Student & Teacher & Target $\tau$ & \bl{first@\,$\tau$} (Base $\rightarrow$ Ours) & Speedup \\
\midrule
Classification (ImageNet) & Too weak & ResNet-50 & MobileNetV3-S & Top-1 65 & \bl{19 ep $\rightarrow$ 37 ep} & \bl{0.51}$\times$ \\
Classification (ImageNet) & Too strong & MobileNetV2 & ResNet-50 & Top-1 65 & 60 ep $\rightarrow$ 57 ep & 1.05$\times$ \\
Classification (ImageNet) & Suitably weaker & ResNet-50 & ResNet-18 & Top-1 65 & \bl{19 ep $\rightarrow$ 4 ep} & \textbf{\bl{4.75}$\times$} \\
\midrule
Detection (COCO) & Too weak & RetinaNet-R50 & RetinaNet-R34 (\bl{earlier} ckpt, AP50$\approx$12.8) & AP50 20.00\% & 10 ep $\rightarrow$ 9 ep & 1.11$\times$ \\
Detection (COCO) & Too strong & RetinaNet-R50 & RetinaNet-R34 (\bl{later} ckpt, AP50$\approx$26.2) & AP50 20.00\% & 10 ep $\rightarrow$ 9 ep & 1.11$\times$ \\
Detection (COCO) & Suitably weaker & RetinaNet-R50 & RetinaNet-R34 (\bl{mid} ckpt, AP50$\approx$19.5) & AP50 20.00\% & 10 ep $\rightarrow$ 6 ep & \textbf{1.67$\times$} \\
\midrule
Generation (CIFAR-10) & Too weak & nc128-rb3 & nc64-rb1 & FID 60 & 16k $\rightarrow$ 20k & 0.80$\times$ \\
Generation (CIFAR-10) & Too strong & nc128-rb3 & nc64-rb3 & FID 60 & 16k $\rightarrow$ 16k & 1.00$\times$ \\
Generation (CIFAR-10) & Suitably weaker & nc128-rb3 & nc64-rb2 & FID 60 & 16k $\rightarrow$ 6k & \textbf{2.67$\times$} \\
\bottomrule
\end{tabular}}
\endgroup
\end{table*}
\vspace{-1.5mm}

\paragraph{\textbf{Teacher Mismatch.}}
As reported in Figure~\ref{fig:teacher_range_v2} and Table~\ref{tab:mismatch_behavior_v2}, the same pattern appears across classification, detection, and generation: suitably weaker teachers give the strongest acceleration, too-weak teachers can remove or reverse speedup \bl{in some settings}, and stronger teachers usually \bl{shrink} gains to near parity.
We next provide mechanism-level evidence for this behavior in Section~\ref{sec:analysis_mismatch_v2}.

\subsubsection{Sensitivity to Optimizers}
\label{sec:exp_v2_ablation_optimizer}

Figure~\ref{fig:ablation_optimizer_v2} evaluates the sensitivity of the proposed method to different optimizers, including SGD, AdamW, and Muon with the same teacher-student pair (R18$\rightarrow$R50)~\cite{He2016ResNet}. 

\subsubsection{Comparison to Label Smoothing}
\label{sec:exp_v2_ablation_ls}

Figure~\ref{fig:ls_kl_ablation_v2}(a) compares label smoothing settings on ImageNet~\cite{Deng2009ImageNet} with the same student-teacher pair and SGD training \bl{recipe}. Our method keeps a clear early-stage advantage under label smoothing, showing that the acceleration effect does not depend on one specific smoothing value. Final Top-1 remains close across settings while \bl{first@\,$\tau$} is still reached earlier by our method.

\subsubsection{Effect of KL Direction }
\label{sec:exp_v2_ablation_kl_direction}

\begin{figure}[t]
  \centering
  \subfloat[Label smoothing ablation on ImageNet.]{\includegraphics[width=0.495\linewidth]{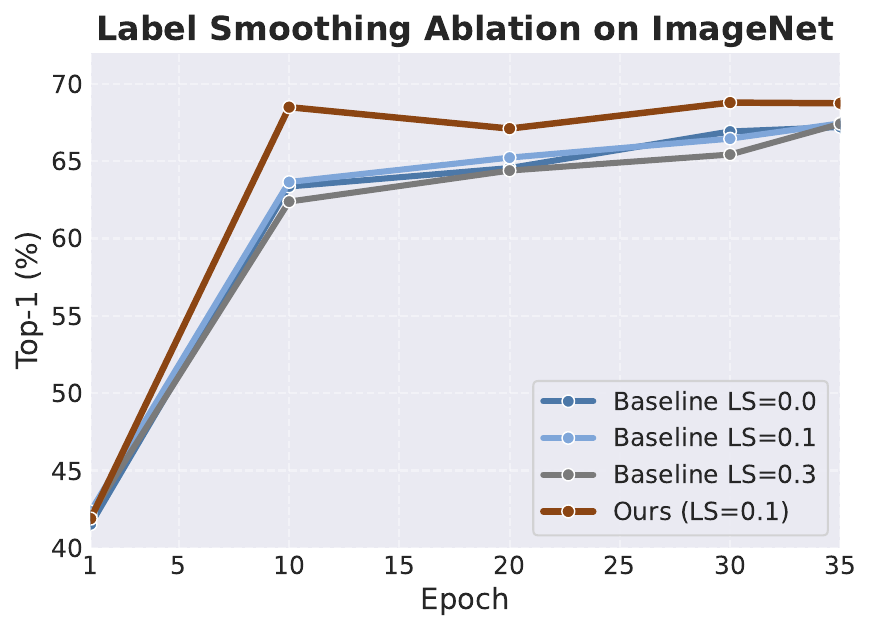}\label{fig:ls_ablation_v2}}\hfill
  \subfloat[KL direction ablation on ImageNet.]{\includegraphics[width=0.495\linewidth]{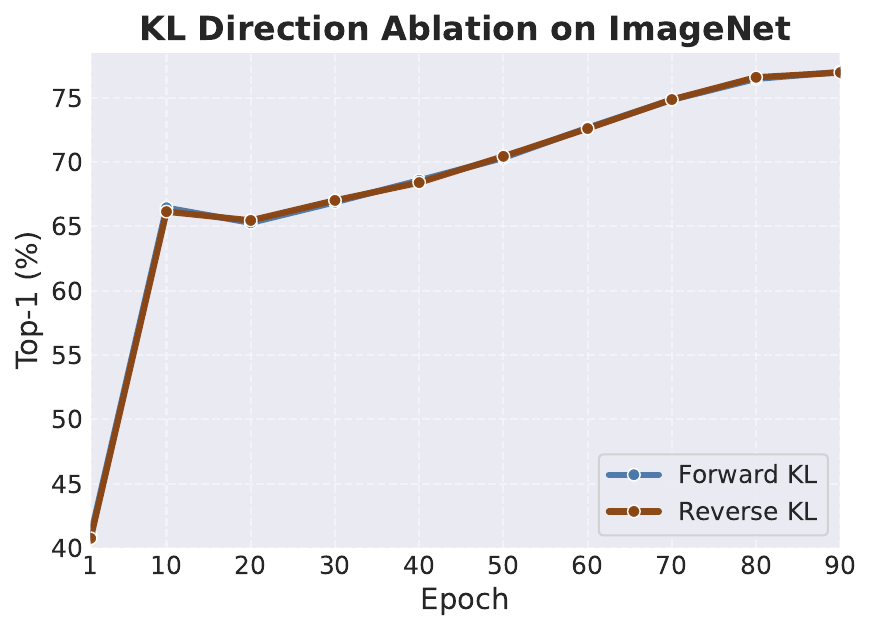}\label{fig:kl_direction_v2}}
  \caption{\textbf{Comparison to Label Smoothing and Evaluation of KL Direction.}
  (a) Our method keeps early-stage acceleration under different label smoothing settings while final Top-1 remains close. (b) Forward and reverse KL variants yield similar trajectories under the same Muon setup, \bl{and the two curves almost completely overlap}.}
  \label{fig:ls_kl_ablation_v2}
\end{figure}

Figure~\ref{fig:ls_kl_ablation_v2}(b) compares forward and reverse KL variants under the same Muon setup~\cite{Jordan2024Muon}. The two trajectories largely overlap, indicating that our acceleration trend is not sensitive to KL direction in this setting. This supports the conclusion that the main gain comes from the weak-to-strong schedule itself rather than from a particular choice of KL direction.

\vspace{-1.5mm}

\subsubsection{Gradient Norms in Early Training}
\label{sec:exp_v2_ablation_gradnorm}

\begin{figure}[t]
  \centering
  \begin{minipage}[t]{0.495\linewidth}
    \vspace{0pt}
    \centering
    \subfloat[Optimizer ablation (fixed teacher-student setup).]{\includegraphics[width=\linewidth]{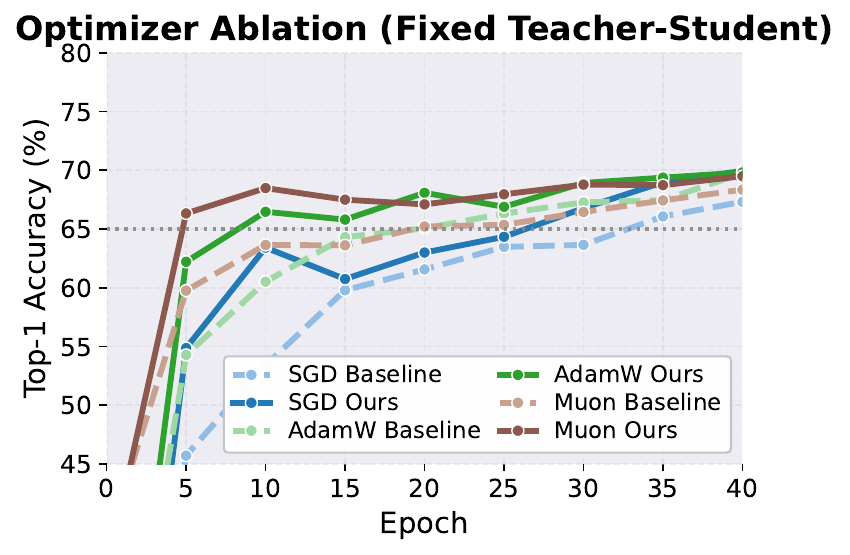}}
  \end{minipage}\hfill
  \begin{minipage}[t]{0.495\linewidth}
    \vspace{0pt}
    \centering
    \subfloat[Gradient norms during early ImageNet training.]{\includegraphics[width=\linewidth]{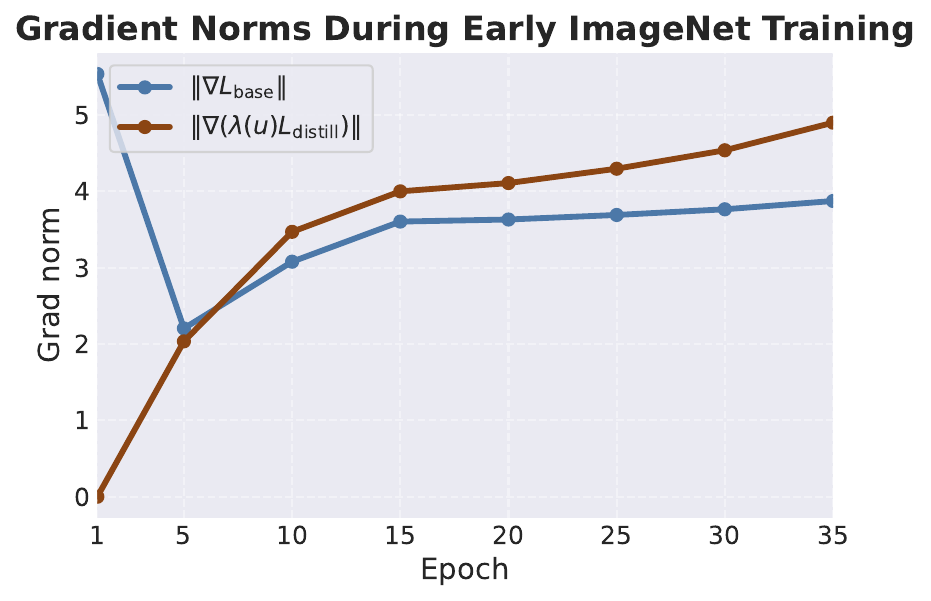}}
  \end{minipage}
  \caption{\textbf{Sensitivity to Optimizer and Optimization Diagnostics.}
  (a) ImageNet optimizer comparison under the same R18$\rightarrow$R50 setup. (b) Early-training gradient norms of the supervised term and distillation term are tracked over epochs; $\|\nabla L_{\text{base}}\|$ and $\|\nabla(\lambda(u)L_{\text{distill}})\|$ stay at comparable magnitudes, indicating gains are not from trivial gradient-rescaling but from the teacher guidance/schedule itself.}
  \label{fig:ablation_optimizer_v2}
  \label{fig:ablation_gradnorm_panel_v2}
  \label{fig:convergence_gradnorm_v2}
\end{figure}

Figure~\ref{fig:ablation_gradnorm_panel_v2} plots gradient norms in the first 35 epochs for the two objective terms used in our method: $L_{\text{base}}$ and $\lambda(u)L_{\text{distill}}$. The $L_{\text{base}}$ coefficient is constant, and the distillation term is scaled only by the scheduled weight $\lambda(u)$. The two gradient components stay in a comparable range and do not show unstable spikes in early training. This indicates that early-stage acceleration is achieved with stable optimization dynamics rather than gradient explosion.

\vspace{-1.5mm}
\begin{figure}[b]
  \centering
  \subfloat[Warm-Start vs no warm-start.]{\includegraphics[height=3.7cm]{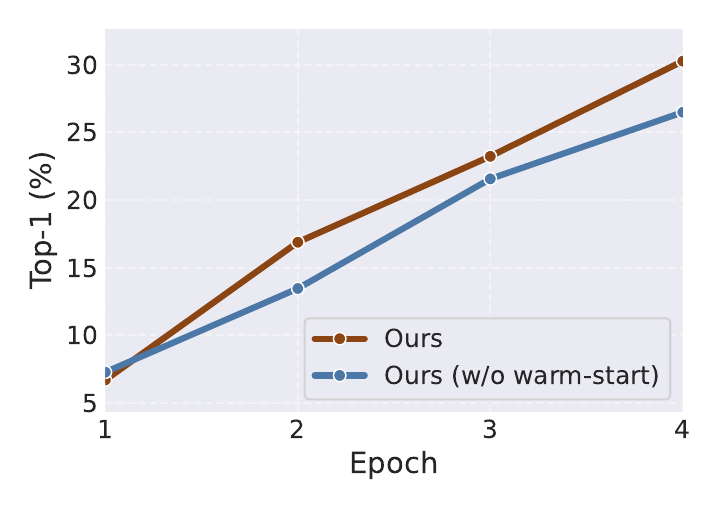}\label{fig:warmstart_ablation_v2}}\hfill
  \subfloat[Post-Surpass Behavior.]{\includegraphics[height=3.7cm]{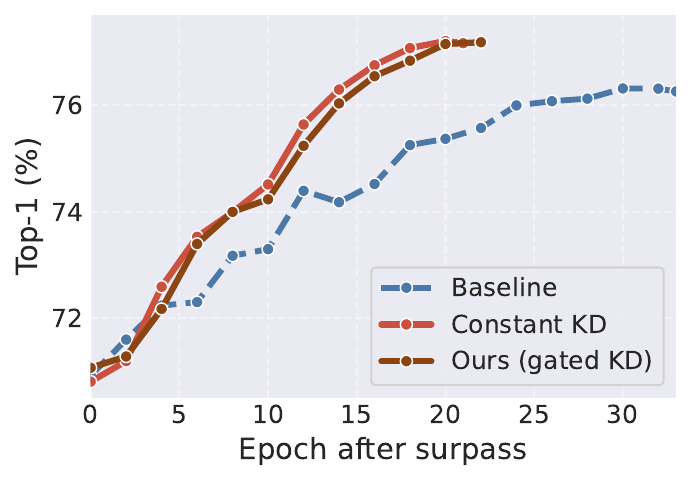}\label{fig:postsurpass_ablation_v2}}
  \caption{\textbf{Ablation of \bl{Warm-Start} and \bl{Stop-After-Surpass}.}
  (a) Warm-start improves early-stage stability and target-reaching; (b) stopping distillation after surpass avoids stale supervision in late training.}
  \label{fig:ablation_warm_stop_v2}
\end{figure}
\vspace{-1.5mm}
\subsubsection{Warm-Start and Stop-After-Surpass}
\label{sec:exp_v2_ablation_warm_stop}

We isolate the two key design choices in our method with two additional ablation plots in Figure~\ref{fig:ablation_warm_stop_v2}. Figure~\ref{fig:ablation_warm_stop_v2}(a) confirms that warm-start gives a smoother and stronger early rise than no warm-start under the same teacher-student pair. 
Figure~\ref{fig:ablation_warm_stop_v2}(b) confirms that once the student has surpassed the teacher, gating distillation off avoids stale late-stage constraints.

\section{Analysis: Teacher Operating Band}
\label{sec:analysis_mismatch_v2}

In this section, we analyze the operating band of the proposed weak teachers.
\label{sec:analysis_teacher_stage_v2}

\paragraph{\textbf{Analysis Setup.}} We analyze three ImageNet~\cite{Deng2009ImageNet} teacher-strength regimes that match the main mismatch discussion: too weak (MobileNetV3-S teacher to ResNet-50 student), suitably weaker (ResNet-18 to ResNet-50), and too strong (ResNet-50 to MobileNetV2)~\cite{Sandler2018MobileNetV2,Howard2019MobileNetV3,He2016ResNet}. For each regime, we keep the same student architecture and training recipe as in the corresponding baseline-versus-ours comparison, and only compare the supervision effect.

\begin{figure*}[t]
  \centering
  \includegraphics[width=0.94\linewidth]{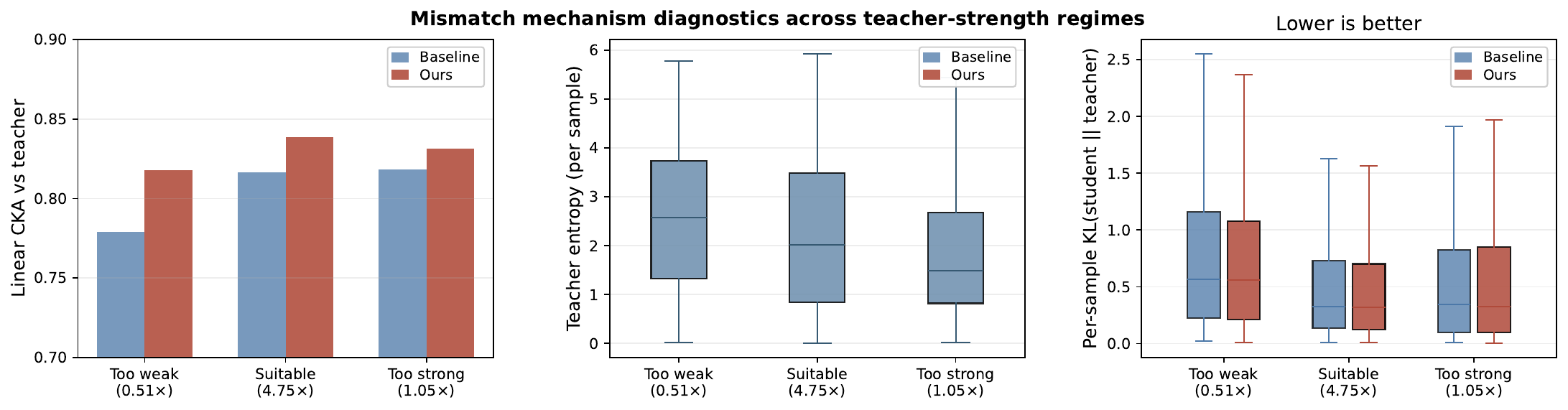}
  \caption{\textbf{Diagnostics for Teacher Mismatch.}
  Left: CKA (student vs.\ teacher features), where higher is better. Middle: teacher entropy, where high means uncertain/ambiguous targets and low means over-specified targets. Right: KL(student$\parallel$teacher), where smaller is better within the \emph{same regime}. Importantly, absolute KL values across different regimes are not directly comparable because teacher distributions differ; the key signal is baseline$\rightarrow$ours change within each regime. We find that too-weak teachers have very high entropy: targets are easy to imitate but not informative enough. Too-strong teachers have very low entropy: targets are informative but too hard for a weaker student to follow early. Suitably-weaker teachers lie in between, giving targets that are both informative and learnable, which matches the largest speedup.}
  \label{fig:mismatch_mechanism_v2}
\end{figure*}
\vspace{-1.5mm}

\paragraph{\textbf{Too Weak Teachers.}} In Figure~\ref{fig:mismatch_mechanism_v2}, the too-weak regime has the highest teacher entropy (mean 2.574), while alignment still improves (CKA 0.426$\rightarrow$0.464, KL 0.817$\rightarrow$0.760). \bl{Here, `improves' means the student KL to \emph{that same teacher} is lower than its baseline in the same regime (so the student is more aligned with that teacher), not that this regime has the lowest KL across all regimes.} Together with the \bl{0.51}$\times$ speedup in Table~\ref{tab:mismatch_behavior_v2}, this indicates that better matching to a weak teacher can still leave acceleration limited because the teacher signal is not decisive enough. The failure mode here is signal quality: alignment improves, but the supervision itself is under-informative.

\paragraph{\textbf{Too Strong Teachers.}}
In the too-strong regime, teacher entropy is lowest (mean 1.910), and alignment gains are also smallest (CKA 0.643$\rightarrow$0.653, KL 0.597$\rightarrow$0.579). This matches the near-parity speedup (1.05$\times$ in Table~\ref{tab:mismatch_behavior_v2}) and suggests a different failure mode: supervision is sharp but less teachable for the weaker student, so additional alignment margin is small. The bottleneck here is teachability under capacity mismatch rather than lack of confidence.

\paragraph{\textbf{Suitably Weaker Teachers.}} The suitably-weaker regime lies between the two extremes in entropy (mean 2.230) while keeping clear alignment gains (CKA 0.591$\rightarrow$0.618, KL 0.527$\rightarrow$0.506). This balanced profile corresponds to the strongest acceleration (4.75$\times$ in Table~\ref{tab:mismatch_behavior_v2}). Therefore, these diagnostics support a two-factor view: acceleration is strongest when teacher signal is both informative enough and teachable enough.

\paragraph{\textbf{Identical Teacher Architecture, Different Checkpoints}.}
For detection on the COCO dataset in Table~\ref{tab:mismatch_behavior_v2}, we intentionally fix the same architecture pair (RetinaNet-R34 teacher, RetinaNet-R50 student) and vary only which training stage \bl{checkpoint} of the same teacher family is used. \bl{Earlier}-stage R34 checkpoints (AP50$\approx$12.8) are too weak and give only marginal gain at AP50@20; \bl{later}-stage R34 checkpoints (AP50$\approx$26.2) are too strong and also collapse to near-parity speedup; \bl{mid}-stage R34 checkpoints (AP50$\approx$19.5) are suitably weaker and recover clear acceleration (1.67$\times$). In Figure~\ref{fig:mismatch_mechanism_v2}, (i) CKA shows whether student features move toward teacher features, (ii) Entropy shows whether teacher targets are too uncertain or too sharp, and (iii) KL shows whether output distributions become easier to match. With the \bl{earlier} checkpoint, alignment improves but entropy is too high, so guidance is weak; with the \bl{later} checkpoint, entropy is very low and alignment gains are small, so guidance is hard to learn from; with the \bl{mid} checkpoint, entropy is moderate and both alignment metrics improve, matching the strongest speedup. This isolates teacher \bl{strength/stage} as the key factor and shows the mismatch effect is not caused by changing teacher architecture.
\section{Discussion}
\label{sec:discussion_v2}

\paragraph{Limitations.}
\bl{We assume that a weak teacher is already available in practical training settings. When teacher training must be included, end-to-end cost depends on how that teacher is obtained and reused across runs; we also report full wall-clock time in supplementary material.

Acceleration is sensitive to teacher--student matching. As reported in our analysis, teachers that are too weak or too strong can reduce gains to near parity. In practice, this means our method benefits from a teacher-selection step, and the best teacher range may vary by task, model family, and optimizer.}
\paragraph{Multi-stage Training.}
\bl{Our training recipe is plug-and-play and can be applied beyond the setting in this paper: (i) model scaling within a family (e.g., small-to-large variants) where a weaker model is already available, (ii) distillation/fine-tuning of larger pretrained foundation models where reducing early optimization time is critical, and (iii) multi-stage training pipelines where teacher outputs can be precomputed once and reused across student runs. A practical next step is an automatic teacher-stage selection and adaptive stop rules so the method can self-tune to different datasets, objectives, and compute budgets.}

\section{Conclusion}
\label{sec:conclusion_v2}

We investigate weak-to-strong distillation as a training-acceleration method rather than a compression method. We propose a plug-and-\bl{play} recipe: use a weaker frozen teacher, apply distillation only in early training, and stop distillation after the student surpasses teacher-level performance.

Across all tested visual learning tasks, including image classification, object detection, and generation, this early-only distillation strategy consistently reduces \bl{first@\,$\tau$} epochs/steps under matched student training setups.
For classification we observe up to 4.8$\times$ speedup, for object detection about 1.7$\times$, and for diffusion generation 2.5$\times$. We analyze why teacher strength matters: we find suitably weaker teachers provide stronger and more teachable guidance than too-weak or too-strong mismatched teachers.

As such, we introduce a plug-and-play training speedup mechanism that is compatible with standard training pipelines and transferable across tasks.

\clearpage 
{
    \small
    \bibliographystyle{splncs04}
    \bibliography{main}
}

\end{document}